\documentclass[11pt,letterpaper]{article}
\usepackage{ijcnlp2017}
\usepackage{times}
\usepackage{latexsym}

\ijcnlpfinalcopy

\def\ijcnlppaperid{***}

\usepackage{bm}
\usepackage{graphicx}
\usepackage{enumitem}
\usepackage{float}
\usepackage{mathtools}
\usepackage{amsfonts}
\usepackage{amsmath}
\usepackage{enumitem}
\usepackage{algorithm}
\usepackage{algorithmic}

\title{Deriving Verb Predicates By Clustering Verbs with Arguments}

\author{Jo\~{a}o Sedoc\textsuperscript{\textdagger} \And Derry Wijaya\textsuperscript{\textdagger} \\\textsuperscript{\textdagger}University of Pennsylvania \And
        Masoud Rouhizadeh\textsuperscript{\textdagger} \And Andy Schwartz\textsuperscript{\textasteriskcentered} \\ \textsuperscript{\textasteriskcentered}Stony Brook University \And Lyle Ungar\textsuperscript{\textdagger}
 \\}

\date{}
\begin{document}
\maketitle
\begin{abstract}
Hand-built verb clusters such as the widely used Levin classes \citep{levin1993english} have proved useful, but have limited coverage.   Verb classes automatically induced from corpus data such as those from VerbKB \citep{wijaya2016verbkb}, on the other hand, can give clusters with much larger coverage, and can be adapted to specific corpora such as Twitter. We present a method for clustering the outputs of VerbKB: verbs with their multiple argument types, e.g.``marry(person, person)", ``feel(person, emotion)." We make use of a novel low-dimensional embedding of verbs and their arguments to produce high quality clusters in which the same verb can be in different clusters depending on its argument type. The resulting verb clusters do a better job than hand-built clusters of predicting sarcasm, sentiment, and locus of control in tweets.
\end{abstract}

\section{Introduction}


English verbs are limited in number (Levin's classes, for instance, include almost 3,100 verbs) and highly polysemous. Depending on its argument realization, a verb may have different semantics or senses \cite{rappaport1998building}. Therefore, including the verb arguments and their semantic types in the semantic analysis should help with sense disambiguation of verbs and their arguments, especially the subject and object. Indeed, verb selectional preferences: the tendencies of verbs to selectively co-occur with specific types of arguments e.g., the verb ``eat" usually takes a type of food as an object argument -- have been shown to be strong indicators of verb diathesis alternations \citep{mccarthy2001lexical}. Furthermore, these selectional preferences can be assigned to the majority of Levin verb classes in VerbNet \citep{schuler2005verbnet}. In this paper we show that clustering verbs along with their subject and object types yields better verb clusters. Verbs are 'disambiguated', such that the same verb ends up in different clusters based on its argument types. Our verb clusters reflect the distribution of verb arguments in  social media language, and provide useful features for modeling this language. 

We propose a method of clustering the governing verbs and their arguments, including the subject, object, and the prepositional phrase. We use as a baseline, Levin's verb classes and  propose new methods for distributional categorization of verbs and their arguments. Unlike Levin's verb classes, our categorization is not limited to verbs; we generate semantic categorization of verbs and their arguments.

A wealth of studies have explored the relation between linguistic features in social media and human traits. However, most studies have used open-vocabulary or bag-of-word approach and few have focused on taking the role of syntactic/semantic contexts and verb argument structure into account. In this study, we show that the verb predicates that we derive improve performance when used as features in models predicting attributes of Facebook messages and Tweets.  Specifically, we look at predicting sarcasm, sentiment, and locus of control: whether the author feels in control or being controlled by the other people. While sarcasm and sentiment are more widely studied, locus of control is a relatively novel task. Our clustering method in effect disambiguates verbs (a highly ambiguous part of speech), and groups together similar verbs by making using of their argument structure.  We show that our automatically derived verb clusters help more in these three prediction tasks than alternatives such as the Levin's classes.

In summary, our main contributions are:
\begin{itemize} [noitemsep,topsep=0pt]
\item we present a novel method for learning the low-dimensional embeddings of verbs and their arguments that takes into account the verb selectional preferences and distribution (section \ref{data:typedverbembeddings}) 
\item we present an algorithm for clustering verbs and their arguments based on the embeddings (section \ref{proposedmethod})
\item we show that our verb clusters outperform hand-built verb classes when used as features for predicting control, sarcasm, and sentiment in tweets (section \ref{clusteringresults})
\end{itemize}






\section{Related Work}


Our approach draws on two different strands of prior work: verb clustering and verb embedding.

\paragraph{Verb Clustering} 
Verb clusters have proved useful for a variety of NLP tasks and applications including e.g., metaphor detection \citep{shutova2010metaphor}, semantic role labeling \citep{palmer2010semantic}, language acquisition \citep{hartshorne2016psych}, and information extraction \citep{nakashole2016machine}. Verb classes are useful because they support generalization and abstraction. VerbNet \citep{schuler2005verbnet} is a widely-used hand-built verb classification which lists over 6,00 verbs that are categorized into 280 classes. The classification is based on Levin's verb classification \citep{levin1993english}, which is motivated by the hypothesis that 
verbs taking similar diathesis alternations tend to share the same meaning and are organized into semantically coherent classes. 
Hand-crafted verb classifications however, suffer from low coverage. This problem has been addressed by various methods to automatically induce verb clusters from corpus data \citep{sun2009improving, 
nakashole2012patty, kawahara2014step, fader2011identifying}. Most recent release is VerbKB \citep{wijaya2016verbkb, wijaya2016mapping}, which contains large-scale verb clusters automatically induced from ClueWeb \citep{callan2009clueweb09}. Unlike previous approaches, VerbKB induces clusters of {\em typed verbs}: verbs (+ prepositions) whose subjects and objects are semantically typed with categories in NELL knowledge base \citep{carlson2010toward} e.g., ``marry on(person, date)", ``marry(person, person)". 

VerbKB clusters 65,000 verbs (+prepositions) and outperforms other large-scale verb clustering methods in terms of how well its clusters align to hand-built verb classes. Unlike these previous works which evaluate the quality of the verb clusters based on their similarities to hand-built verb classes, we evaluate our verb clusters directly against hand-built verb classes (Levin, VerbNet) on their utility in building predictive models for assessing control, sarcasm, and sentiment. 

\paragraph{Verb Embeddings}
Word embeddings are vector space models that represent  words  as  real-valued vectors in a low-dimensional semantic space based on their contexts in large corpora. Recent approaches for learning these vectors such as word2vec \citep{mikolov2013distributed} and Glove \citep{pennington2014glove} are widely used. However, these models represent each word with a single unique vector. Since verbs are highly polysemous, individual verb senses should potentially each have their own embeddings. Sense-aware word embeddings such as \citep{reisinger2010multi,huang2012improving,neelakantan2015efficient,li2015multi} can be useful
. However, they base their representations solely on distributional statistics obtained from corpora, ignoring semantic roles or types of the verb arguments. Recent study by \citet{schwartz2016symmetric} has observed that verbs are different than other parts of speech in that their distributional representation can benefit from taking verb argument role into accounts. These argument roles or types can be provided by existing semantic resources. However, learning sense-aware embeddings that take into account information from existing semantic resources \citep{iacobacci2015sensembed} requires large amounts of sense-annotated corpora. Since we have only data in the form of (subject, verb, object) triples extracted from ClueWeb, the limited context\footnote{window size of 1, limited syntactic information, and no sentence or whole document context} also means that traditional word embedding models or word sense disambiguation systems may not learn well on the data \citep{melamud2016role}. 

Motivated by previous works that have shown verb selectional preferences to be useful for verb clustering \citep{sun2009improving, wijaya2016verbkb} and that verb 
distributional representation can benefit from taking into account the verb argument roles \citep{schwartz2016symmetric}, we cluster VerbKB typed verbs by first learning novel, low-dimensional representations of the\textit{ typed verbs}, thus encoding information about the verb selectional preferences and distribution in the data.

We learn embeddings of typed verbs (verbs plus the type of their subjects and 
objects) in VerbKB. 
Unlike traditional one-word-one-vector embedding, we learn embeddings for each typed verb e.g., the embedding for ``abandon(person, person)" is separate from the embedding for ``abandon(person, religion)". 
Using only triples in the form of (subject, verb, object) extracted from ClueWeb, we learn verb embeddings by treating each verb as a relation between its subject and object \citep{bordes2013translating}. 
Since verbs are predicates that express relations between the arguments and adjuncts in sentences, we believe this is a natural way for representing verbs. 

We cluster typed verbs based on their embeddings. Then, at run time, given any text containing a verb and its arguments, we straightforwardly map the text to the verb clusters by assigning types to the verb arguments using NELL's noun phrase to category mapping\footnote{publicly available at \url{http://rtw.ml.cmu.edu/rtw/nps}} to obtain the typed verb and hence, its corresponding verb clusters. This differs from sense-aware embedding approaches that require the text at run time to be sense-disambiguated with the learned senses, a difficult problem by itself.  

\section{Method}
\label{proposedmethod}

\begin{figure*}[tbh!]
    \centering
    \includegraphics[width=\textwidth]{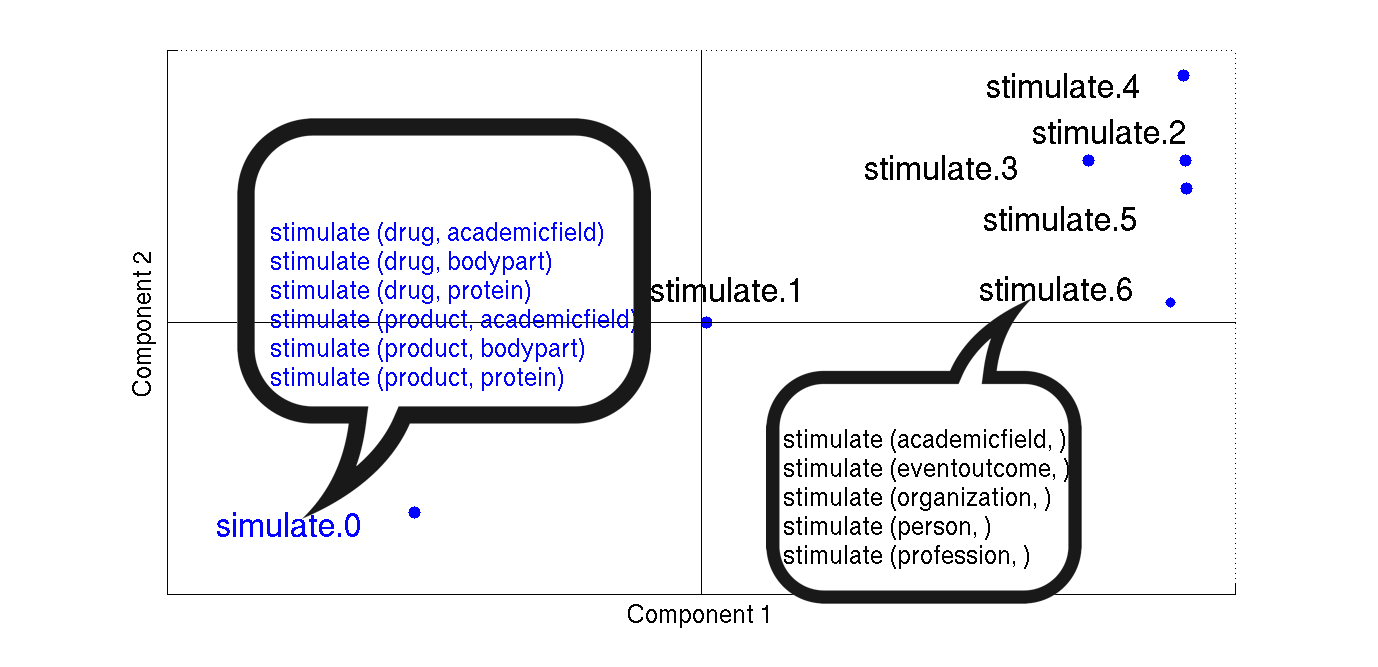}
    \caption{After algorithm~\ref{algo:cluster_algo} of the clustering algorithm, the different argument types of each verb are clustered. For example, the verb ``stimulate'' here has 6 clusters (The number of clusters came from the number of WordNet senses for the verb ``stimulate".)} 
    \label{fig:stimulate_clusters}
\end{figure*}

\begin{figure}[tbh!]
    \centering
    \includegraphics[width=0.5\textwidth]{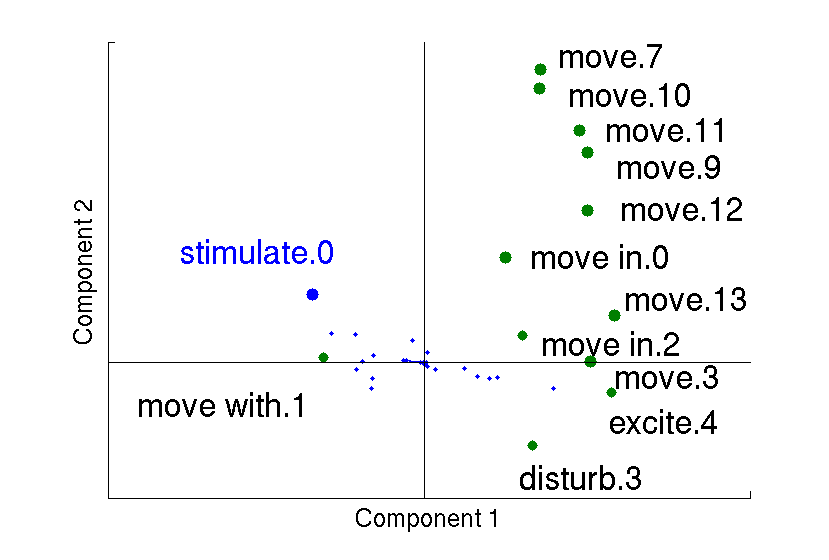}
    \caption{The final output of the clustering algorithm~\ref{algo:cluster_algo2} is the clusters of verb senses. 
    This example cluster shows one sense of the verb ``stimulate": ``stimulate.0'' which is clustered with different senses of ``move''. The small points represent additional words groups in the cluster which are not displayed.}
    \label{fig:example_cluster}
\end{figure}

Given the embeddings of the typed verbs, the main goal of our clustering is to create representations of verbs using their argument structure similar in concept to the hand curated Levin classes, but with higher coverage and precision. Our method comprises four steps:
\begin{itemize}[noitemsep,topsep=0pt]
\item shallow parsing the sentence into subject, verb (+ preposition), and object
\item labeling the subject and object into their NELL categories
\item  identifying the clustering within each verb (+ preposition) as in figure~\ref{fig:stimulate_clusters}
\item indexing into the cluster of between verb cluster embeddings as shown in figure~\ref{fig:example_cluster}. 
\end{itemize}

We use algorithm 1 for creating verbal argument clusters for \textit{each} verb, and  algorithm 2 to cluster \textit{between} the verbal argument clusters. 
This process results in verb predicate clusters with are conceptually similar to Levin class, but which include prepositions as well as arguments and are in practice closer to VerbNet and FrameNet classes.

\paragraph{ Step 1: Parsing and lemmatization}
The first step in our pipeline for labeling the verb predicate is to parse the sentence or tweet (detailed in section \ref{tweetprocessing}). 
Then, we extracted the words in in the nominal subject, direct object position, and the prepositional phrases 
and reduced morphological variations by lemmatizing the verbs and their arguments. This whole process captured the sentence kernel.

\paragraph{ Step 2: Subject and object NELL categorization}
Subsequently, the subject and object noun phrases are mapped to NELL categories. This categorization creates an abstract view of the verbal arguments into types. 

\paragraph{ Step 3: Verb-specific verb argument clusters}
In order to create verb (+ preposition) argument clusters for each verb, all typed embeddings 
for the verb are clustered using spectral clustering method of \citet{yu2003multiclass} for multiclass normalized cuts.  The number of clusters is limited to the WordNet 3.1 \citep{miller1995wordnet} senses for each verb. The centers of the 
clusters are the representative embedding for the cluster. One can interpret these clusters as ``synsets'' of verbal arguments which are similar in embedding space. This created a mapping $f$ from the verb with its preposition $v$, the subject NELL category $s$, and the object category $o$ to the verb arguments cluster and the cluster's representative embedding.

\begin{algorithm}[H]
   \caption[Cluster Algo]{Verb-Specific Argument Clustering Algorithm}
\label{algo:cluster_algo}
\begin{algorithmic}[1]
   \STATE {\bfseries Input:} Embeddings, $emb(v, t_s, t_o)$, for a set of typed verbs containing the verb (+ preposition) $v$, its subject type $t_s$ and object type $t_o$
   \FOR{Each verb (+ preposition) $v$ over all arguments, $emb(v, *, *)$}
     \STATE Set $k^v$ to the number of word senses from WordNet 3.1 \citep{miller1995wordnet} with a default of 2 for missing verbs.
     \STATE Calculate the affinity matrix $W^{sim}$ using a cosine similarity between each embedding from $emb(v,*,*)$.
     \STATE Find $k^v$ clusters ($C_i^v$) from $W^{sim}$.
     \STATE Keep a map from $f$ from verb $v$, subject type $t_s$, and object type $t_o$ to the cluster number $C_i^v$.
     \STATE Calculate the mean of the embeddings $e_{C_i^v}$.
   \ENDFOR
   \STATE {\bfseries Output:} The verb sense embeddings $[e_{C_i^v}]$ for all verbs, the mapping function $f$.
\end{algorithmic}
\end{algorithm}
\noindent
The main output from algorithm~\ref{algo:cluster_algo} are verb argument clusters and embeddings $e_{f(v,t_s,t_o)}$
. These clusters can be considered as verb ``sense'' clusters.
In figure~\ref{fig:stimulate_clusters} we showed the $e_{C^{simulate}}$ plotted with respect to the first and second principle components in the 
verb sense embedding space. ``stimulate.0'' is further from the rest of the 
verb sense embeddings for ``stimulate''.

\paragraph{ Step 4: Clustering 
between verb argument clusters} 
The final component in the procedure is to cluster across verb argument clusters i.e., ``verb senses" using the clusters' representative embeddings. Here we also include side thesaurus information in order to maintain semantic similarity particularly by including antonym information. We follow the procedure of \citet{sedoc2016semantic} which extends spectral clustering to account for negative edges. 

\begin{algorithm}[H]
   \caption[Cluster Algo]{Verb Predicate Clustering Algorithm}
\label{algo:cluster_algo2}
\begin{algorithmic}[1]
   \STATE {\bfseries Input:} Cluster embeddings from Algorithm~\ref{algo:cluster_algo} $[e_{C_i^v}]$, the thesaurus, $T$, and the number of clusters $k$.
   \STATE Calculate the 
   verb 
   senses affinity matrix $W$ using the radial basis function of the Euclidean distance between $e_{C_i^v}$ and $e_{C_{j}^{v'}}$.
   \STATE Find $k$ clusters $\mathcal{C}$ using signed spectral clustering of $W$ and $T$.
   \STATE Keep a function $g$ from $C_i^v$ to the cluster number $\mathcal{C}_j$
\STATE {\bfseries Output:} The verb sense embeddings $[e_{C_i^v}]$, the mapping function $f$, and $g$.
\end{algorithmic}
\end{algorithm}
\noindent
The main result having run algorithm~\ref{algo:cluster_algo2} are verb predicate clusters of typed verbs $(v,t_s,t_o)$ from $g( f(v,t_s,t_o) )$.

Figure \ref{fig:example_cluster} corresponds to a verb predicate cluster 
which includes ``stimulate.0'' but not other senses of ``stimulate''. Furthermore, ``stimulate.0'' is grouped with various senses of ``move''. This shows how the two step clustering algorithm is effective in creating clusters which are similar in purpose to Levin classes. 

\section{Prediction tasks}

We use the verb predicate clusters as features in three prediction tasks: estimating locus of control, sarcasm, and sentiment from social media language. We now briefly describe these three tasks and the data set we use for them.

\subsection{Locus of control} Locus of control, or "control," is defined as the degree to which a person is in control of others or situation or being controlled by them. A large number of studies explored the role of control (or locus of control, LoC) on the physical and mental health. They have found that a person's perceived LoC can influence their health \cite{lachman1998sense},  well-being \cite{krause1984stress}, and career prospects \cite{judge2002measures}. All of these studies are limited to small populations (mainly based on questionnaires) and none of them propose automated large-scale methods

We deployed a survey on Qualtrics, comprising several demographic questions as well as a set of 128 items, and invited users to share access to their Facebook status updates. 2465 subjects reported their age, gender and items indicative of their general health and well-being. We split each Facebook status update into multiple sentences and asked three trained annotators to determine for each sentences if the author is in control (internal control) or being controlled by others or circumstances (external control). The inter-annotator agreement between the three annotators was around \%76. We took the majority vote of the annotator for each message and assigned binary labels for internal and external control. 

\subsection{Sarcasm} 
Several number of studies have used surface linguistic features \cite{carvalho2009clues,davidov2010semi}, language patterns \cite{davidov2010semi}, lexical features and emotions \cite{gonzalez2011identifying}, counter-factuals, unexpectedness, emotions, and n-grams \cite{reyes2013multidimensional}. Other works have explored the role of social context in detecting sarcasm as well \cite{rajadesingan2015sarcasm,bamman2015contextualized}.
 \citet{schifanella2016detecting} worked on multimodal sarcasm analysis and detection.
Our method advances on predicting sarcasm using word embeddings \citep{ghosh2015sarcastic,joshi2016automatic} to verb predicates.

Here we use the dataset from \citet{bamman2015contextualized} including 17,000 tweets. The tweets are semi-automaticaly  annotated for sarcasm (e.g. using \#sarcasm). The dataset contains 51\% sarcastic and 49\% non-sarcastic manually annotated tweets (not likely to reflect of real-world rates of sarcastic tweets). 

\subsection{Sentiment} Sentiment has been extremely widely studied~\citep{pang2008opinion,liu2012survey}. Both surface level as well as lexical structure have been shown to be useful in the task of sentiment prediction~\citep{neviarouskaya2009semantically}. Large corpora are available, both at the document level as well as the tweet level where sentiment has been assessed. In our work, we used the sentiment prediction task to compare verb predicate clusters with hand-curated verb classes on this task.

\section{Data preprocessing}

\subsection{Social media text corpus} Our corpus for verb clustering consists of the status updates of 15,000 Facebook users, a subset of the ones who volunteered to share their posts in the ``MyPersonality'' application \cite{kosinski2013private}, between January 2009 and October 2011. The users had English as a primary language and were less than 65 years old (due to data sparsity beyond this age).

\subsection{Data processing and extracting verb arguments}
\label{tweetprocessing}
We first perform a text normalization pipeline that cleans each tweet or Facebook status update  (removes emoticon, URLs, email addresses, handles, hashtags, etc.), does spelling correction and partial abbreviation expansion, and reduces the number of repeated characters. Then, we tokenize and split Facebook status updates into sentences (we keep tweets as single sentences). We tokenize the tweets using CMU ARK Twitter Twokenize script \citep{owoputi2013improved,o2010tweetmotif}. 
Next, we obtained dependency parses of our corpus using SyntaxNet with Parsey McParseface model\footnote{\url{https://github.com/tensorflow/models/tree/master/syntaxnet}} that provides universal dependencies in ({\em relation, head, dependent}) triples\footnote{In our in-house evaluation SyntaxNet with Parsey McParseface model outperformed Stanford Parser \cite{socher2013parsing} on social media domain and it is essentially better than the Tweebo Parser \citep{kong2014dependency} that does not provide dependency relations}.  
%
We extracted subject, verb, object, preposition and the object of preposition from the dependency trees, lemmatizing each word using NLTK wordNet lemmatizer \cite{bird2009natural}.
Given the nature of twitter data the parses of the tweets are very noisy and created errors, such as, ``rying('t.t', None)'' from {\it ``I've planted my ca t.t rying to grow cat tails for Halloween .''} Nonetheless, parsing twitter is out of scope for this paper and we used the same parse for all methods. 

\subsection{Typed verb embeddings}
\label{data:typedverbembeddings}
Typed verbs in VerbKB \citep{wijaya2016verbkb} are created by extracting subject, verb (lemmatized), object, preposition and the object of preposition from the dependency trees in the ClueWeb corpus\citep{callan2009clueweb09}. Triples in the form of (subject, verb (+preposition), object) are extracted, and the subjects and objects are typed using the NELL knowledge base categories \citep{carlson2010toward}. The type signatures of verbs e.g., (person, person) for ``marry" are then selected based on their frequencies of occurrence in the corpus using Resnik's selectional association scores \citep{resnik1997selectional}. The result is a collection of triples of typed verbs with their subject and object noun phrases (NPs) in ClueWeb e.g., (Barack\_Obama, marry(person, person), Michelle\_Obama), (Tom\_Hanks, marry(person, person), Rita\_Wilson). 

Inspired by \citet{bordes2013translating}, who model relationships by interpreting them as translations operating on the low-dimensional embeddings of the entities, we learn low-dimensional representations of the typed verbs by interpreting them as translations operating on the low-dimensional embeddings of their subject and object noun phrases. Specifically, given a set of triples: ($n_s$, $v_t$, $n_o$) composed of the subject and object NP $n_s$, $n_o$ $\in N$ (the set of NPs) and the typed verb $v_t$, we want the embedding of the object NP $\bm{n_o}$ to be a nearest neighbor of $\bm{n_s} + \bm{v_t}$ i.e., $\bm{n_s} + \bm{v_t} \approx \bm{n_o}$ when ($n_s$, $v_t$, $n_o$) is observed in ClueWeb and far away otherwise. Using $L_2$ distance $d$, following \citet{bordes2013translating}, to learn the embeddings we minimize over the set $S$ of triples observed in ClueWeb:

{\small
\begin{align*}
\mathcal{L} = \sum_{(n_s,v_t,n_o) \in S} \sum_{(n_s',v_t,n_o') \in S'} [\gamma & + d(\bm{n_s} + \bm{v_t}, \bm{n_o}) \\
& - d(\bm{n_s'}+\bm{v_t},\bm{n_o'})]_{+}
\end{align*}}
where $[x]_{+}$ denotes the positive part of $x$, $\gamma > 0$ is a hyperparameter and $S'$ is the set of corrupted triples constructed as in \citet{bordes2013translating}. 

For typed intransitives (e.g., ``sleep(person)"), since they do not have object NPs, we learn their embeddings by making use of their prepositions and objects e.g., ``sleep in(person, location)" whose triples are observed in ClueWeb. Specifically, given triples in the form of ($v_i$, $p$, $n_o$) composed of the intransitive verb $v_i$, the preposition $p$ and the preposition object NP $n_o$ e.g., (sleep(person), in, adjacent\_room), we want the embeddings to be $\bm{v_i} + \bm{p} \approx \bm{n_o}$ when ($v_i$, $p$, $n_o$) is observed in ClueWeb and far away otherwise.

We use a fast implementation \citep{lin2015learning} of \citet{bordes2013translating} to learn 300 dimensional embeddings for transitive and intransitive typed verbs using this approach with 100 epochs. We use the implementation's default setting for other parameters. 


\subsection{GloVe Embedding}
As a baseline, we used the 200 dimensional word embeddings from \citet{pennington2014glove}~\footnote{\url{http://nlp.stanford.edu/projects/glove/}}. trained using Wikipedia 2014 + Gigaword 5 (6B tokens). GloVe has been shown to have better correlation with semantic relations than Word2Vec Skip-Gram embeddings from \citet{mikolov2013distributed} \citep{pennington2014glove}.

\section{Clustering Results}
\label{clusteringresults}
\paragraph{Baselines}
We used several baselines for clustering. Levin classes are split into several forms. We used the most fine-grained classes, which clusters verbs into 199 categories. GloVe clusters were created using K-means clustering. 
The clustering was done 
by averaging the subject, verb, and object vectors

\paragraph{Verb Predicate Clusters}

We took a subset of VerbKB typed verb 
embeddings from the extracted vocabulary of 15,000 parsed Facebook posts as well as our control, sarcasm, and sentiment data. From the vocabulary of Levin verbs, verbs from Facebook status updates with subject, verb, object that occur more than twice, and verbs from Twitter sentiment and control data, we obtain 6,747 verbs. This is subsequently intersected with the VerbKB typed verbs 
vocabulary of 46,960 verbs with prepositions attached, which results in 3791 verbs (+prepositions) 
Finally, once arguments are added the vocabulary expands to 322,564 
typed verbs which are clustered according to algorithm~\ref{algo:cluster_algo} and algorithm ~\ref{algo:cluster_algo2} to yield the final verb predicate clusters. 

Table \ref{tab:emotion} shows an example of different verb senses that have the same object type, which are clustered in the same verb predicate cluster. 

\begin{table}[t]
    \centering
    \begin{tabular}{l|c|c}
    verb & subject & object \\ \hline
clarify & jobposition & emotion \\
erode & event & emotion \\
lose & personcanada & emotion \\
deny & writer & emotion \\
lament & athlete & emotion \\
exploit & jobposition & emotion \\
fidget & person & emotion \\ 
prove & celebrity & emotion \\ 
raise & filmfestival & emotion \\ 
make & militaryconflict & emotion \\

    \end{tabular}
    \caption{This is a subset of the verb predicate cluster that has emotion as object.}
    \label{tab:emotion}
\end{table}

Table \ref{tab:beat} shows various verb predicate clusters of the verb ``beat'', which is particularly interesting for predicting control.  For example, ``The Patriots beat the Falcons.'', ``I beat John with a stick.'', and ``My father beat me.'', will all have different measures of control. 

\begin{table}[H]
    \centering
    \begin{tabular}{l|c|c|c}
    verb & subject & object & cluster \# \\ \hline
beat & personus & person & 138 \\
beat & personasia & person & 138 \\
beat & personmexico & person & 138 \\
beat & personus & athlete & 195 \\
beat & personcanada & athlete & 195 \\
beat & coach & organization & 195 \\
    \end{tabular}
    \caption{There are multiple senses of ``beat'' which are shown to be in different clusters. Cluster number 138 includes ``hit'' and ``crash''. Whereas, ``block'', ``run'', and ``win'' are members of cluster 195. }
    \label{tab:beat}
\end{table}

\section{Results and Discussion}

We perform a set of experiments to extrinsically evaluate verb predicate clusters. As baselines we use Levin classes, VerbNet, as well as clusters of subject, verb, object GloVe embeddings. In order to evaluate the verb predicate clusters, we used the clustering method to make various clusters using both transitive as well as intransitive verb clusters. 

The results from table~\ref{tab:results} show that our verb predicate clusters 
outperform Levin classes, VerbNet categories, as well as clusters of GloVe vector averaging the subject, verb and object (S-V-O clusters). We also tried other baselines, including logistic regression of GloVe embeddings instead of clustering and the results where F-score of 0.657, 0.612, and 0.798 for control, sarcasm, and sentiment respectively.  We also tried to change the number of clusters to 200 to match the fine grained Levin classes. 

\begin{table}[tbh!]
    \centering
    \begin{tabular}{l|c|c|c}
 &	control	& sarcasm &	sentiment \\ \hline
Levin &	0.660	& 0.619	& 0.804 \\
VerbNet	& 0.679	& 0.628	& 0.796 \\
S-V-O clusters &	0.685	& 0.621	& 0.795 \\
Verb Predicate &	{\bf 0.721}	& {\bf 0.637}	& {\bf 0.807 }
    \end{tabular}
    \caption{Comparison of the F-score of the Levin classes, VerbNet, GloVe embedding clusters and our verb predicate clusters for predicting control, sentiment, and sarcasm of tweets. Ten fold cross-validation was used on the datasets.}
    \label{tab:results}
\end{table}

One shortfall of typed verb embeddings is due to the poor coverage for common nouns in NELL KB. In order to alleviate this issue we tried creating a manual list of the most frequent common nouns in our dataset to NELL categories. Unfortunately, this problem is systemic and only a union with something akin to WordNet would suffice to solve this issue. For instance, the sense of ``root'' is categorized with ``poke'', ``forage'', ``snoop'', ``rummage'' and others in this sense; however, the sense as well as all of the afore mentioned words aside from ``root'' are not covered by type verb embedding. This is definitely an avenue of improvement which should be explored in the future.

\section{Conclusion}

Verb predicates are  empirically driven clusters which disambiguate both verb sense as well as synonym set. Verbal predicates were shown to outperform Levin classes,  in predicting control, sarcasm, and sentiment. These verbal predicates are similar to Levin classes in spirit while having increased precision and coverage.

For future work, we intend to integrate social media data in to build better verb arguments clusters, i.e. clusters that help with better prediction.

\bibliography{verb_predicate}

\begin{thebibliography}{}
\expandafter\ifx\csname natexlab\endcsname\relax\def\natexlab#1{#1}\fi

\bibitem[{Bamman and Smith(2015)}]{bamman2015contextualized}
David Bamman and Noah~A Smith. 2015.
\newblock Contextualized sarcasm detection on twitter.
\newblock In {\em ICWSM\/}. Citeseer, pages 574--577.

\bibitem[{Bird et~al.(2009)Bird, Klein, and Loper}]{bird2009natural}
Steven Bird, Ewan Klein, and Edward Loper. 2009.
\newblock {\em Natural language processing with Python: analyzing text with the
  natural language toolkit\/}.
\newblock " O'Reilly Media, Inc.".

\bibitem[{Bordes et~al.(2013)Bordes, Usunier, Garcia-Duran, Weston, and
  Yakhnenko}]{bordes2013translating}
Antoine Bordes, Nicolas Usunier, Alberto Garcia-Duran, Jason Weston, and Oksana
  Yakhnenko. 2013.
\newblock Translating embeddings for modeling multi-relational data.
\newblock In {\em Advances in neural information processing systems\/}. pages
  2787--2795.

\bibitem[{Callan et~al.(2009)Callan, Hoy, Yoo, and Zhao}]{callan2009clueweb09}
Jamie Callan, Mark Hoy, Changkuk Yoo, and Le~Zhao. 2009.
\newblock Clueweb09 data set.

\bibitem[{Carlson et~al.(2010)Carlson, Betteridge, Kisiel, Settles,
  Hruschka~Jr, and Mitchell}]{carlson2010toward}
Andrew Carlson, Justin Betteridge, Bryan Kisiel, Burr Settles, Estevam~R
  Hruschka~Jr, and Tom~M Mitchell. 2010.
\newblock Toward an architecture for never-ending language learning.
\newblock In {\em AAAI\/}. volume~5, page~3.

\bibitem[{Carvalho et~al.(2009)Carvalho, Sarmento, Silva, and
  De~Oliveira}]{carvalho2009clues}
Paula Carvalho, Lu{\'\i}s Sarmento, M{\'a}rio~J Silva, and Eug{\'e}nio
  De~Oliveira. 2009.
\newblock Clues for detecting irony in user-generated contents: oh...!! it's so
  easy;-.
\newblock In {\em Proceedings of the 1st international CIKM workshop on
  Topic-sentiment analysis for mass opinion\/}. ACM, pages 53--56.

\bibitem[{Davidov et~al.(2010)Davidov, Tsur, and Rappoport}]{davidov2010semi}
Dmitry Davidov, Oren Tsur, and Ari Rappoport. 2010.
\newblock Semi-supervised recognition of sarcastic sentences in twitter and
  amazon.
\newblock In {\em Proceedings of the fourteenth conference on computational
  natural language learning\/}. Association for Computational Linguistics,
  pages 107--116.

\bibitem[{Fader et~al.(2011)Fader, Soderland, and
  Etzioni}]{fader2011identifying}
Anthony Fader, Stephen Soderland, and Oren Etzioni. 2011.
\newblock Identifying relations for open information extraction.
\newblock In {\em Proceedings of the Conference on Empirical Methods in Natural
  Language Processing\/}. Association for Computational Linguistics, pages
  1535--1545.

\bibitem[{Ghosh et~al.(2015)Ghosh, Guo, and Muresan}]{ghosh2015sarcastic}
Debanjan Ghosh, Weiwei Guo, and Smaranda Muresan. 2015.
\newblock Sarcastic or not: Word embeddings to predict the literal or sarcastic
  meaning of words.
\newblock In {\em EMNLP\/}. pages 1003--1012.

\bibitem[{Gonz{\'a}lez-Ib{\'a}nez et~al.(2011)Gonz{\'a}lez-Ib{\'a}nez, Muresan,
  and Wacholder}]{gonzalez2011identifying}
Roberto Gonz{\'a}lez-Ib{\'a}nez, Smaranda Muresan, and Nina Wacholder. 2011.
\newblock Identifying sarcasm in twitter: a closer look.
\newblock In {\em Proceedings of the 49th Annual Meeting of the Association for
  Computational Linguistics: Human Language Technologies: Short Papers-Volume
  2\/}. Association for Computational Linguistics, pages 581--586.

\bibitem[{Hartshorne et~al.(2016)Hartshorne, O’Donnell, Sudo, Uruwashi, Lee,
  and Snedeker}]{hartshorne2016psych}
Joshua~K Hartshorne, Timothy~J O’Donnell, Yasutada Sudo, Miki Uruwashi,
  Miseon Lee, and Jesse Snedeker. 2016.
\newblock Psych verbs, the linking problem, and the acquisition of language.
\newblock {\em Cognition\/} 157:268--288.

\bibitem[{Huang et~al.(2012)Huang, Socher, Manning, and
  Ng}]{huang2012improving}
Eric~H Huang, Richard Socher, Christopher~D Manning, and Andrew~Y Ng. 2012.
\newblock Improving word representations via global context and multiple word
  prototypes.
\newblock In {\em Proceedings of the 50th Annual Meeting of the Association for
  Computational Linguistics: Long Papers-Volume 1\/}. Association for
  Computational Linguistics, pages 873--882.

\bibitem[{Iacobacci et~al.(2015)Iacobacci, Pilehvar, and
  Navigli}]{iacobacci2015sensembed}
Ignacio Iacobacci, Mohammad~Taher Pilehvar, and Roberto Navigli. 2015.
\newblock Sensembed: Learning sense embeddings for word and relational
  similarity.
\newblock In {\em Proceedings of the 53th Annual Meeting of the Association for
  Computational Linguistics\/}. Association for Computational Linguistics,
  pages 95--105.

\bibitem[{Joshi et~al.(2016)Joshi, Bhattacharyya, and
  Carman}]{joshi2016automatic}
Aditya Joshi, Pushpak Bhattacharyya, and Mark~James Carman. 2016.
\newblock Automatic sarcasm detection: A survey.
\newblock {\em arXiv preprint arXiv:1602.03426\/} .

\bibitem[{Judge et~al.(2002)Judge, Erez, Bono, and
  Thoresen}]{judge2002measures}
Timothy~A Judge, Amir Erez, Joyce~E Bono, and Carl~J Thoresen. 2002.
\newblock Are measures of self-esteem, neuroticism, locus of control, and
  generalized self-efficacy indicators of a common core construct?

\bibitem[{Kawahara et~al.(2014)Kawahara, Peterson, and
  Palmer}]{kawahara2014step}
Daisuke Kawahara, Daniel Peterson, and Martha Palmer. 2014.
\newblock A step-wise usage-based method for inducing polysemy-aware verb
  classes.
\newblock In {\em ACL (1)\/}. pages 1030--1040.

\bibitem[{Kong et~al.(2014)Kong, Schneider, Swayamdipta, Bhatia, Dyer, and
  Smith}]{kong2014dependency}
Lingpeng Kong, Nathan Schneider, Swabha Swayamdipta, Archna Bhatia, Chris Dyer,
  and Noah~A Smith. 2014.
\newblock A dependency parser for tweets .

\bibitem[{Kosinski et~al.(2013)Kosinski, Stillwell, and
  Graepel}]{kosinski2013private}
Michal Kosinski, David Stillwell, and Thore Graepel. 2013.
\newblock Private traits and attributes are predictable from digital records of
  human behavior.
\newblock {\em Proceedings of the National Academy of Sciences\/}
  110(15):5802--5805.

\bibitem[{Krause and Stryker(1984)}]{krause1984stress}
Neal Krause and Sheldon Stryker. 1984.
\newblock Stress and well-being: The buffering role of locus of control
  beliefs.
\newblock {\em Social Science \& Medicine\/} 18(9):783--790.

\bibitem[{Lachman and Weaver(1998)}]{lachman1998sense}
Margie~E Lachman and Suzanne~L Weaver. 1998.
\newblock The sense of control as a moderator of social class differences in
  health and well-being.
\newblock {\em Journal of personality and social psychology\/} 74(3):763.

\bibitem[{Levin(1993)}]{levin1993english}
Beth Levin. 1993.
\newblock {\em English verb classes and alternations: A preliminary
  investigation\/}.
\newblock University of Chicago press.

\bibitem[{Li and Jurafsky(2015)}]{li2015multi}
Jiwei Li and Dan Jurafsky. 2015.
\newblock Do multi-sense embeddings improve natural language understanding?
\newblock In {\em Proceedings of the 2015 Conference on Empirical Methods in
  Natural Language Processing\/}. Association for Computational Linguistics,
  pages 1722--1732.

\bibitem[{Lin et~al.(2015)Lin, Liu, Sun, Liu, and Zhu}]{lin2015learning}
Yankai Lin, Zhiyuan Liu, Maosong Sun, Yang Liu, and Xuan Zhu. 2015.
\newblock Learning entity and relation embeddings for knowledge graph
  completion.
\newblock In {\em Twenty-Ninth AAAI Conference on Artificial Intelligence\/}.

\bibitem[{Liu and Zhang(2012)}]{liu2012survey}
Bing Liu and Lei Zhang. 2012.
\newblock A survey of opinion mining and sentiment analysis.
\newblock In {\em Mining text data\/}, Springer, pages 415--463.

\bibitem[{McCarthy(2001)}]{mccarthy2001lexical}
Diana McCarthy. 2001.
\newblock {\em Lexical acquisition at the syntax-semantics interface: diathesis
  alternations, subcategorization frames and selectional preferences.\/}.
\newblock Ph.D. thesis, University of Sussex.

\bibitem[{Melamud et~al.(2016)Melamud, McClosky, Patwardhan, and
  Bansal}]{melamud2016role}
Oren Melamud, David McClosky, Siddharth Patwardhan, and Mohit Bansal. 2016.
\newblock The role of context types and dimensionality in learning word
  embeddings.
\newblock In {\em Proceedings of NAACL-HLT 2016\/}. Association for
  Computational Linguistics, pages 1030--1040.

\bibitem[{Mikolov et~al.(2013)Mikolov, Sutskever, Chen, Corrado, and
  Dean}]{mikolov2013distributed}
Tomas Mikolov, Ilya Sutskever, Kai Chen, Greg~S Corrado, and Jeff Dean. 2013.
\newblock Distributed representations of words and phrases and their
  compositionality.
\newblock In {\em Advances in neural information processing systems\/}. pages
  3111--3119.

\bibitem[{Miller(1995)}]{miller1995wordnet}
George~A Miller. 1995.
\newblock Wordnet: a lexical database for english.
\newblock {\em Communications of the ACM\/} 38(11):39--41.

\bibitem[{Nakashole and Mitchell(2016)}]{nakashole2016machine}
Ndapandula Nakashole and Tom~M Mitchell. 2016.
\newblock Machine reading with background knowledge.
\newblock {\em arXiv preprint arXiv:1612.05348\/} .

\bibitem[{Nakashole et~al.(2012)Nakashole, Weikum, and
  Suchanek}]{nakashole2012patty}
Ndapandula Nakashole, Gerhard Weikum, and Fabian Suchanek. 2012.
\newblock Patty: a taxonomy of relational patterns with semantic types.
\newblock In {\em Proceedings of the 2012 Joint Conference on Empirical Methods
  in Natural Language Processing and Computational Natural Language
  Learning\/}. Association for Computational Linguistics, pages 1135--1145.

\bibitem[{Neelakantan et~al.(2014)Neelakantan, Shankar, Passos, and
  McCallum}]{neelakantan2015efficient}
Arvind Neelakantan, Jeevan Shankar, Alexandre Passos, and Andrew McCallum.
  2014.
\newblock Efficient non-parametric estimation of multiple embeddings per word
  in vector space.
\newblock In {\em Proceedings of the 2014 Conference on Empirical Methods in
  Natural Language Processing (EMNLP)\/}. Association for Computational
  Linguistics, pages 1059--1069.

\bibitem[{Neviarouskaya et~al.(2009)Neviarouskaya, Prendinger, and
  Ishizuka}]{neviarouskaya2009semantically}
Alena Neviarouskaya, Helmut Prendinger, and Mitsuru Ishizuka. 2009.
\newblock Semantically distinct verb classes involved in sentiment analysis.
\newblock In {\em IADIS AC (1)\/}. pages 27--35.

\bibitem[{O'Connor et~al.(2010)O'Connor, Krieger, and Ahn}]{o2010tweetmotif}
Brendan O'Connor, Michel Krieger, and David Ahn. 2010.
\newblock Tweetmotif: Exploratory search and topic summarization for twitter.
\newblock In {\em ICWSM\/}. pages 384--385.

\bibitem[{Owoputi et~al.(2013)Owoputi, O'Connor, Dyer, Gimpel, Schneider, and
  Smith}]{owoputi2013improved}
Olutobi Owoputi, Brendan O'Connor, Chris Dyer, Kevin Gimpel, Nathan Schneider,
  and Noah~A Smith. 2013.
\newblock Improved part-of-speech tagging for online conversational text with
  word clusters.
\newblock Association for Computational Linguistics.

\bibitem[{Palmer et~al.(2010)Palmer, Gildea, and Xue}]{palmer2010semantic}
Martha Palmer, Daniel Gildea, and Nianwen Xue. 2010.
\newblock Semantic role labeling.
\newblock {\em Synthesis Lectures on Human Language Technologies\/}
  3(1):1--103.

\bibitem[{Pang et~al.(2008)Pang, Lee et~al.}]{pang2008opinion}
Bo~Pang, Lillian Lee, et~al. 2008.
\newblock Opinion mining and sentiment analysis.
\newblock {\em Foundations and Trends{\textregistered} in Information
  Retrieval\/} 2(1--2):1--135.

\bibitem[{Pennington et~al.(2014)Pennington, Socher, and
  Manning}]{pennington2014glove}
Jeffrey Pennington, Richard Socher, and Christopher~D Manning. 2014.
\newblock Glove: Global vectors for word representation.
\newblock In {\em EMNLP\/}. volume~14, pages 1532--1543.

\bibitem[{Rajadesingan et~al.(2015)Rajadesingan, Zafarani, and
  Liu}]{rajadesingan2015sarcasm}
Ashwin Rajadesingan, Reza Zafarani, and Huan Liu. 2015.
\newblock Sarcasm detection on twitter: A behavioral modeling approach.
\newblock In {\em Proceedings of the Eighth ACM International Conference on Web
  Search and Data Mining\/}. ACM, pages 97--106.

\bibitem[{Rappaport~Hovav and Levin(1998)}]{rappaport1998building}
Malka Rappaport~Hovav and Beth Levin. 1998.
\newblock Building verb meanings.
\newblock {\em The projection of arguments: Lexical and compositional
  factors\/} pages 97--134.

\bibitem[{Reisinger and Mooney(2010)}]{reisinger2010multi}
Joseph Reisinger and Raymond~J Mooney. 2010.
\newblock Multi-prototype vector-space models of word meaning.
\newblock In {\em Human Language Technologies: The 2010 Annual Conference of
  the North American Chapter of the Association for Computational
  Linguistics\/}. Association for Computational Linguistics, pages 109--117.

\bibitem[{Resnik(1997)}]{resnik1997selectional}
Philip Resnik. 1997.
\newblock Selectional preference and sense disambiguation.
\newblock In {\em Proceedings of the ACL SIGLEX Workshop on Tagging Text with
  Lexical Semantics: Why, What, and How\/}. Washington, DC, pages 52--57.

\bibitem[{Reyes et~al.(2013)Reyes, Rosso, and
  Veale}]{reyes2013multidimensional}
Antonio Reyes, Paolo Rosso, and Tony Veale. 2013.
\newblock A multidimensional approach for detecting irony in twitter.
\newblock {\em Language resources and evaluation\/} 47(1):239--268.

\bibitem[{Schifanella et~al.(2016)Schifanella, de~Juan, Tetreault, and
  Cao}]{schifanella2016detecting}
Rossano Schifanella, Paloma de~Juan, Joel Tetreault, and Liangliang Cao. 2016.
\newblock Detecting sarcasm in multimodal social platforms.
\newblock In {\em Proceedings of the 2016 ACM on Multimedia Conference\/}. ACM,
  pages 1136--1145.

\bibitem[{Schuler(2005)}]{schuler2005verbnet}
Karin~Kipper Schuler. 2005.
\newblock Verbnet: A broad-coverage, comprehensive verb lexicon .

\bibitem[{Schwartz et~al.(2016)Schwartz, Reichart, and
  Rappoport}]{schwartz2016symmetric}
Roy Schwartz, Roi Reichart, and Ari Rappoport. 2016.
\newblock Symmetric patterns and coordinations: Fast and enhanced
  representations of verbs and adjectives.
\newblock In {\em Proceedings of NAACL-HLT\/}. pages 499--505.

\bibitem[{Sedoc et~al.(2016)Sedoc, Gallier, Ungar, and
  Foster}]{sedoc2016semantic}
Jo{\~a}o Sedoc, Jean Gallier, Lyle Ungar, and Dean Foster. 2016.
\newblock {Semantic Word Clusters Using Signed Normalized Graph Cuts}.
\newblock {\em arXiv preprint arXiv:1601.05403\/} .

\bibitem[{Shutova et~al.(2010)Shutova, Sun, and Korhonen}]{shutova2010metaphor}
Ekaterina Shutova, Lin Sun, and Anna Korhonen. 2010.
\newblock Metaphor identification using verb and noun clustering.
\newblock In {\em Proceedings of the 23rd International Conference on
  Computational Linguistics\/}. Association for Computational Linguistics,
  pages 1002--1010.

\bibitem[{Socher et~al.(2013)Socher, Bauer, Manning, and
  Ng}]{socher2013parsing}
Richard Socher, John Bauer, Christopher~D Manning, and Andrew~Y Ng. 2013.
\newblock Parsing with compositional vector grammars.
\newblock In {\em ACL (1)\/}. pages 455--465.

\bibitem[{Sun and Korhonen(2009)}]{sun2009improving}
Lin Sun and Anna Korhonen. 2009.
\newblock Improving verb clustering with automatically acquired selectional
  preferences.
\newblock In {\em Proceedings of the 2009 Conference on Empirical Methods in
  Natural Language Processing: Volume 2-Volume 2\/}. Association for
  Computational Linguistics, pages 638--647.

\bibitem[{Wijaya(2016)}]{wijaya2016verbkb}
Derry~Tanti Wijaya. 2016.
\newblock {\em VerbKB: A Knowledge Base of Verbs for Natural Language
  Understanding\/}.
\newblock Ph.D. thesis, Carnegie Mellon University.

\bibitem[{Wijaya and Mitchell(2016)}]{wijaya2016mapping}
Derry~Tanti Wijaya and Tom~M Mitchell. 2016.
\newblock Mapping verbs in different languages to knowledge base relations
  using web text as interlingua.
\newblock In {\em Proceedings of NAACL-HLT\/}. pages 818--827.

\bibitem[{Yu and Shi(2003)}]{yu2003multiclass}
Stella~X Yu and Jianbo Shi. 2003.
\newblock Multiclass spectral clustering.
\newblock In {\em Computer Vision, 2003. Proceedings. Ninth IEEE International
  Conference on\/}. IEEE, pages 313--319.

\end{thebibliography}
\bibliographystyle{emnlp_natbib}

\end{document}